# Universe Points Representation Learning for Partial Multi-Graph Matching

**Zhakshylyk Nurlanov**[1,2], **Frank R. Schmidt**[1], **Florian Bernard**[2]

[1] Bosch Center for Artificial Intelligence
[2] University of Bonn
{zhakshylyk.nurlanov, frank.r.schmidt}@de.bosch.com, fb@uni-bonn.de

## Abstract

Many challenges from natural world can be formulated as a graph matching problem. Previous deep learning-based methods mainly consider a full two-graph matching setting. In this work, we study the more general partial matching problem with multi-graph cycle consistency guarantees. Building on a recent progress in deep learning on graphs, we propose a novel data-driven method (URL) for partial multi-graph matching, which uses an object-to-universe formulation and learns latent representations of abstract universe points. The proposed approach advances the state of the art in semantic keypoint matching problem, evaluated on Pascal VOC, CUB, and Willow datasets. Moreover, the set of controlled experiments on a synthetic graph matching dataset demonstrates the scalability of our method to graphs with large number of nodes and its robustness to high partiality.

## 1 Introduction

Many real-world problems of finding correspondences can be formalized as graph matching. The examples of tasks in computer vision and graphics, where graph matching plays a crucial role, are object tracking (Wang et al. 2015), optical flow estimation (Sun et al. 2018), shape matching (Gao et al. 2021) and image keypoint matching (Fey et al. 2020). Graph matching is also relevant in other domains for matching protein networks (Singh, Xu, and Berger 2008), comparing molecules (Kriege, Humbeck, and Koch 2019), or linking users in social networks (Zhang and Philip 2015).

The recent progress in end-to-end learning of graph matching has allowed one to find the most useful graph features instead of hand-crafting them (Zanfir and Sminchisescu 2018), and also enabled the incorporation of axiomatic graph matching solvers into learning models (Rolínek et al. 2020). In this work, we specifically focus on finding image keypoint correspondences across an image collection with the help of deep learning techniques. We consider the general setting assuming that the images in the collection depict different instances of objects from the same category, as illustrated in Fig. 1 for the object category cat. A strong advantage of our multi-matching approach compared to previous pairwise matching formulations (Fig. 1a) is that due to our *object-to-universe matching representation* we obtain cycle-consistent multi-matchings. Moreover, as opposed to the previous template-based deep multi-graph matching approach (Fig. 1b), our object-to-universe matching representation does not require an explicit template shape. Instead, we propose to learn latent representations of abstract universe points through aligning them with graph node features (Fig. 1c, Fig. 3 right). The learned embeddings encode higher-order information with the help of graph neural network processing and allow to find accurate object-to-universe matchings.

In addition, the object-to-universe formulation naturally supports the challenging partial multi-matching setting, as demonstrated in Fig. 1d. This is in stark contrast to a large portion of recent deep graph matching approaches that only tackle the full (bijective) matching problem. However, in a real-world collection of images, not all keypoints may be observed in each image, e.g. due to ambiguities of 2D image projections stemming from occlusions, camera-view change, object deformation, shape variability, etc. Hence, these methods require a keypoint pre-filtering before matching. As such, the partial matching setting is the most natural one and has thus the highest relevance in practical scenarios, since it can be applied to *unfiltered keypoints*.

Since image keypoints represent 2D projections of 3D points, we would like to expand the dimensionality of the matched keypoints. Therefore, we propose to learn a third *virtual* coordinate lifting (Fig. 4), and to incorporate it into the graph neural network architecture (3D SplineCNN (Fey et al. 2018)). The third dimension is learned with respect to the task of graph matching, thus it is a virtual coordinate. As we demonstrate in our ablation study in Sec. 4, the use of 3D coordinates leads to an improved matching performance.

Overall, our method advances the state of the art (SOTA) on various keypoint matching benchmarks. Namely, it achieves a 71.7% F1-score on the Pascal VOC dataset and a 95.1% F1-score on the CUB dataset on partial multi-graph matching without keypoint filtering. Additionally, it gives 81.8% accuracy on Pascal VOC dataset on full two-graph matching with intersection-filtered keypoints. Moreover, we show that the learned node features taken from a model pretrained on (reduced) Pascal VOC can be transferred without fine-tuning to the Willow dataset, yielding 98.9% accuracy. On a set of controlled experiments, we demonstrate the robustness of the proposed approach to the partiality of the input data, its scalability to large graphs, and compu-

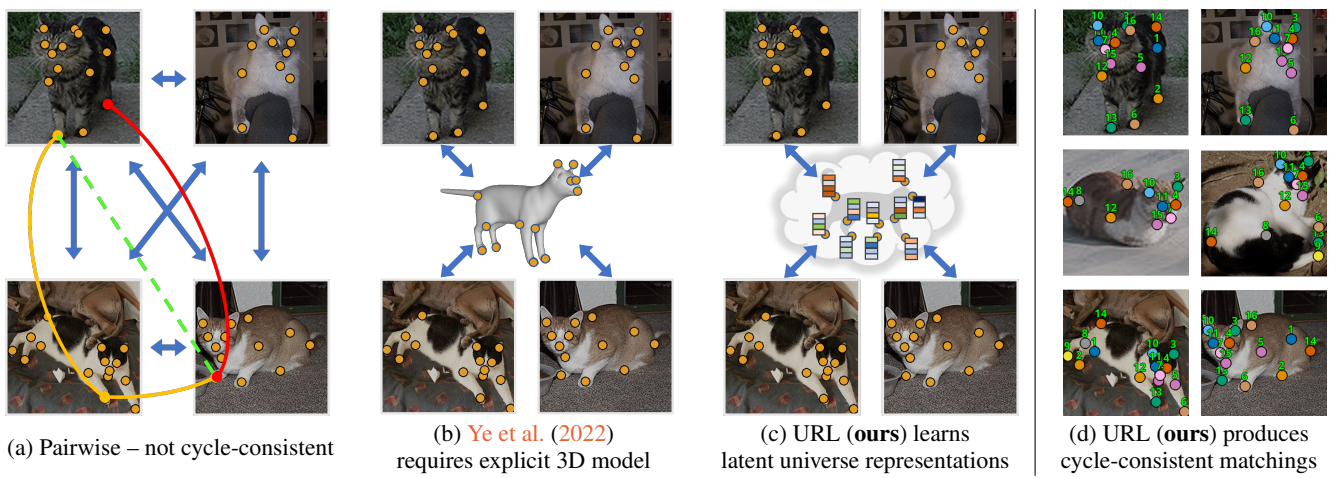

(a) Pairwise – not cycle-consistent  (b) Ye et al. (2022) requires explicit 3D model  (c) URL (**ours**) learns latent universe representations  (d) URL (**ours**) produces cycle-consistent matchings

Figure 1: Overview of pairwise and multi-graph matching methods. **(a)** Pairwise matchings from prior works do not guarantee cycle consistency. The *red* arc shows an incorrect matching that violates the cycle consistency constraint, indicated by the *green dashed* line. **(b)** Previous cycle-consistent approach by Ye et al. (2022) requires an explicit template of sparse 3D geometry of the universe object. **(c)** Our approach learns latent representations of abstract universe points containing geometric and visual information, which allows to produce accurate cycle-consistent matchings under non-rigid deformations and occlusions **(d)**.

tational efficiency compared to the strong baseline BBGM-Multi (Rolínek et al. 2020).

We summarize our main contributions as follows:

- We propose the deep learning approach for *partial* multi-graph matching based on universe points representation learning;
- Building on the recent progress in graph-based learning, we introduce a virtual coordinate lifting, thus improving 2D image keypoint matching performance;
- The proposed method achieves state-of-the-art results in partial multi-graph matching with unfiltered keypoints on Pascal VOC and CUB datasets.

## Related Work

In this section, we introduce prior works on deep graph matching and discuss related methods.

**Deep graph matching** Graph matching consists of two main parts: feature extraction from raw input data and a matching algorithm. The seminal paper (Zanfir and Sminchisescu 2018) on two-graph matching has demonstrated how learnable features can be combined with a differentiable graph matching solver. Recent progress in this area allows differentiating through general combinatorial solvers (Rolínek et al. 2020). Another direction considers solving the simple linear assignment problem (LAP) (Munkres 1957) (via differentiable Sinkhorn iterations (Adams and Zemel 2011)) instead of the quadratic assignment problem QAP (Lawler 1963). The higher-order information is incorporated into the node features via graph neural network processing (Yu et al. 2019). This also allows an iterative improvement of soft matchings (Fey et al. 2020). Another way to solve a graph matching problem is by learning a solver on a product graph (Wang et al. 2020). While these methods allow end-to-end learning, they focus on matching pairs of graphs, and for most of the solvers, the unrealistic setting of full (bijective) matchings is assumed. In contrast, we match multiple graphs in a cycle-consistent way and naturally allow for the realistic setting of partial matching.

**Multi-Graph Matching** In previous deep learning works that addressed multi-graph matching, cycle consistency was obtained via permutation synchronization of predicted pairwise matchings (Wang, Yan, and Yang 2020, 2021). In contrast, when using an object-to-universe matching formulation, cycle consistency can be directly obtained by construction (Schmidt et al. 2007; Huang and Guibas 2013). The deep learning-based approach (Ye et al. 2022) operates in this setting, but explicitly reconstructs a 3D model of a universe graph. While this gives the desired cycle consistency, the method requires 3D reconstruction, and it does not perform on par with the state-of-the-art pairwise matching methods (Rolínek et al. 2020; Wang, Yan, and Yang 2021). In our work, we propose to find an object-to-universe matching without explicitly reconstructing a 3D model of the universe. Instead, we learn latent representations of universe points by matching them with node features.

**Graph Neural Networks** One of the main sources of the recent progress in deep graph matching is the rise of graph neural network (GNN) architectures (Scarselli et al. 2008). Previous works have implemented various GNNs with intra- and cross-graph communication to optimally learn interconnections between nodes (Wang, Yan, and Yang 2019). Particularly, for the problem of keypoint matching, a SplineCNN architecture (Fey et al. 2018, 2020) has shown promising performance (Wang, Yan, and Yang 2021). The reason is that it allows to efficiently incorporate geometric information directly into the node features. Similarly to these methods, we utilize the SplineCNN architecture for our image keypoint matching problem. However, we propose to also predict a third virtual coordinate, which helps to resolve 2D keypoint location conflicts. As we later demonstrate, this leads to an improved matching performance.

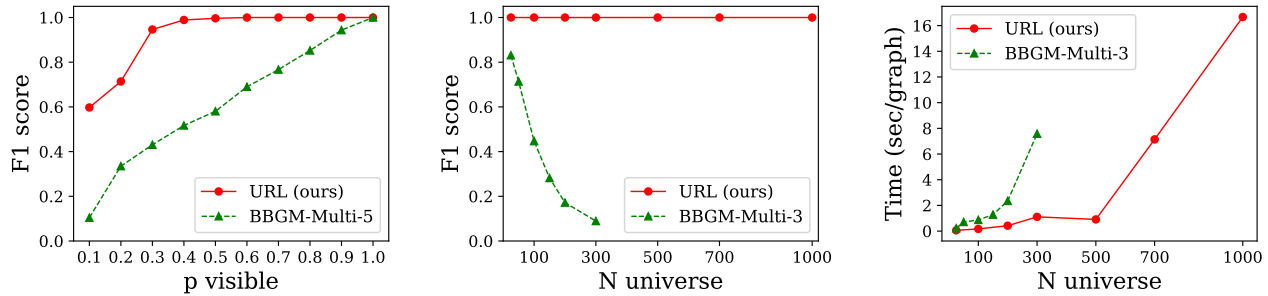

Figure 2: Partial multi-graph matching performance evaluation on synthetic dataset. Our method (URL) handles well **high partiality** in input points (left plot, low visibility). Its accuracy does not degrade on **larger problems** (middle plot, large number of points). And it scales **more efficiently** (right plot, processing time) than previous method BBGM-Multi.

## 2 Background

In the following, we provide the general background for the multi-graph matching problem. First, we introduce the notation concerning graphs and matchings. Then we revisit the theory behind the object-to-universe multi-graph matching.

**Notation**

We assume that $k \in \mathbb{N}$ different graphs $\mathcal{G}_1, \ldots, \mathcal{G}_k$ are given, where for each $i \in [k] := \{1, \ldots, k\}$ the graph $\mathcal{G}_i = (\mathcal{V}_i, \mathcal{E}_i, F_i, E_i)$ comprises of a node set $\mathcal{V}_i$, an edge set $\mathcal{E}_i \subset \mathcal{V}_i \times \mathcal{V}_i$, node features $F_i : \mathcal{V}_i \to \mathcal{F}_1$, and edge features $E_i : \mathcal{E}_i \to \mathcal{F}_2$. The feature spaces $\mathcal{F}_1$ and $\mathcal{F}_2$ are some finite-dimensional vector spaces and we write $m_i = |\mathcal{V}_i|$ for the amount of nodes in $\mathcal{G}_i$.

In this work, we consider the task of *partial* multi-graph matching, in which we want to find correspondences between nodes of the graphs $\mathcal{G}_1, \ldots, \mathcal{G}_k$. Note that the number of nodes $m_i$ can differ among the graphs $\mathcal{G}_i$, i.e. not all of the matched nodes are present in each graph, which backs the term 'partial'. Before we introduce our formalism for multi-graph matching, we introduce the correspondence $X_{ij} \in \mathbb{P}_{m_i m_j}$ between a pair of graphs $(\mathcal{G}_i, \mathcal{G}_j)$, where the set of *partial permutation matrices* is given as

$$\mathbb{P}_{m_i m_j} = \{X \in \{0,1\}^{m_i \times m_j} : \\ X \mathbf{1}_{m_j} \leq \mathbf{1}_{m_i}, \mathbf{1}_{m_i}^\top X \leq \mathbf{1}_{m_j}^\top \}, \quad (1)$$

for $\mathbf{1}_m$ being the $m$-dimensional vector of ones.

**Multi-graph matching**

When considering multiple ($k > 2$) graphs, it is necessary to ensure that the pairwise matchings $\{X_{ij}\}_{i,j=1}^k$ do not contradict each other. Namely, the matchings should be *cycle-consistent* (Bernard et al. 2019a):

**Definition 1** (Cycle consistency). *The set $\mathcal{X} = \{X_{ij}\}_{i,j=1}^k$ of pairwise matchings is said to be cycle-consistent if for all $i, j, \ell \in [k]$ it holds that*

$$X_{ii} = I_{m_i}, \quad \text{(Identity)}$$
$$X_{ij} = X_{ji}^\top, \quad \text{(Symmetry)}$$
$$X_{ij} X_{j\ell} \leq X_{i\ell}. \quad \text{(Partial Transitivity)}$$

The property of (Partial Transitivity) introduces a cubic number of non-convex quadratic constraints. A more efficient way to characterize cycle-consistent matchings is by using the notion of *universe matchings*.

Assuming that we are given an upper bound of the *universe size* $d \in \mathbb{N}$, for graph $\mathcal{G}$ with $m$ nodes ($m \leq d$) the space of *object-to-universe* matchings is denoted as

$$\mathbb{U}_{md} = \{X \in \{0,1\}^{m \times d} : X \mathbf{1}_d = \mathbf{1}_m, \mathbf{1}_m^\top X \leq \mathbf{1}_d^\top \}. \quad (2)$$

Then the cycle-consistency constraints can be reformulated in the following manner (Huang and Guibas 2013):

**Lemma 1** (Universe matchings). *The set $\mathcal{X} = \{X_{ij}\}_{i,j=1}^k$ of pairwise matchings is cycle-consistent, iff there exists a universe of size $d$ and a collection of object-to-universe matchings $\{X_i \in \mathbb{U}_{m_i d}\}_{i=1}^k$, such that for each $X_{ij} \in \mathcal{X}$ it holds that $X_{ij} = X_i X_j^T$.*

In other words, by matching each node of a graph with exactly one abstract *universe point*, we can find object-to-universe matchings $\{X_i \in \mathbb{U}_{m_i d}\}_{i=1}^k$, which by construction ensures the cycle consistency of pairwise matchings.

## 3 Proposed Method

Below, we introduce our method for partial multi-graph matching by **u**niverse points **r**epresentation **l**earning (URL). We first formulate the general problem setup and clarify the assumptions of our method. Then we formalize the proposed approach as a deep learning problem and introduce our proposed neural network architecture with a novel virtual coordinate lifting block.

**Problem Setup and Assumptions**

Our method accepts as input a set of graphs $\{\mathcal{G}_i\}_{i=1}^k$ that are to be matched. It outputs globally cycle-consistent pairwise matchings from object-to-universe matchings, i.e. $\{\hat{X}_{ij} := \hat{X}_i \hat{X}_j^\top \in \mathbb{P}_{m_i m_j}\}_{i=1}^k$.

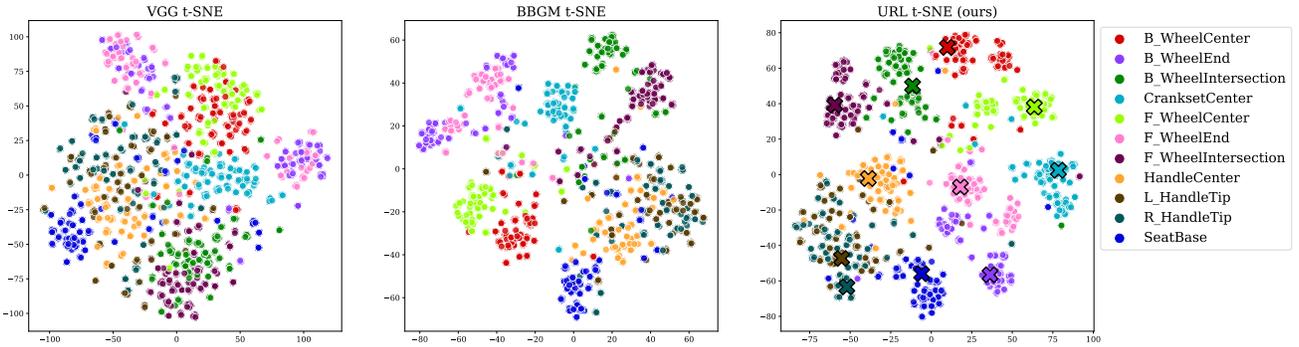

Figure 3: Two dimensional t-SNE embeddings of latent keypoint representations from different neural networks (on `bicycle` object class from Pascal VOC dataset). Both BBGM (middle) and URL (right, ours) models start with the node features from the ImageNet pre-trained VGG network (left), and further improve them. We empirically demonstrate that matching keypoints with learnable latent representations of *universe points* (centroids ✖) yields more accurate results than matching keypoints with each other (see Table 1, BBGM vs URL).

**Object classes** Since the set of matched graphs $\{\mathcal{G}_i\}_{i=1}^k$ share common nodes, we expect that the graphs correspond to the same abstract class, called object class. In other words, each graph represents an object from the same object class. We *assume* that the input graphs are from a known object class during inference. Therefore, we consider matchings for different object classes independently.

**Supervision** Further, we *require* the ground truth matchings for the training set as a supervisory signal. We note that any set of correct pairwise matchings can be transformed into object-to-universe matching form via matrix factorization (Bernard et al. 2019a). Therefore, we can always extract the ground truth universe matchings $\{X_i^{\text{gt}}\}_{i=1}^k$ from pairwise matchings $\{X_{ij}^{\text{gt}}\}_{i,j=1}^k$ during training. From this, we *estimate* an upper bound of the number of universe points $d$.

The above-stated assumptions hold for multi-image keypoint matching problems with predefined sets of detected keypoints sharing similar semantic meanings. For example, commonly used datasets, such as Pascal VOC, Willow, CUB, provide keypoint annotations as node labels, i.e. in object-to-universe matching form, and give the ground truth number of universe points $d$.

## Deep Object-to-Universe Matching

Given the problem setup, we propose to utilize the object-to-universe matching formulation from Lemma 1. This allows to match a set of graphs $\{\mathcal{G}_i\}_{i=1}^k$ with abstract universe points individually. Thus, it reduces the memory and time consumption compared to multi-graph matching approaches based on synchronization of all pairwise matchings.

To facilitate the data-driven approach, we propose to simultaneously learn the feature encoder of graph nodes and the latent representations of universe points. Formally, given a graph $\mathcal{G}_i$ with its initial node and edge features $F_i$ and $E_i$, we update them to produce new node and edge features $\hat{F}_i$ and $\hat{E}_i$. This is achieved with a parameterized graph-processing function $\Psi_\theta$, where $\theta$ denotes the learnable parameters:

$$\Psi_\theta(\mathcal{G}_i) =: \hat{\mathcal{G}}_i = (\mathcal{V}_i, \mathcal{E}_i, \hat{F}_i, \hat{E}_i) \qquad (3)$$

In practice, $\Psi_\theta$ is implemented as a graph neural network. As a consequence, the node features contain the neighborhood (including edge) information. Thus, we will only use the improved node features $\hat{F}_i \in \mathbb{R}^{m_i \times h}$ for matching. The features are from $h$-dimensional vector space.

Concurrently, we learn the latent representations of $d$ universe points $\hat{U} \in \mathbb{R}^{d \times h}$ in the same $h$-dimensional vector space. The *soft* (non-binary, continuous) object-to-universe matching is then found by the normalization of feature similarities as

$$\hat{S}_i := \text{Norm}(\hat{F}_i \hat{U}^\top) \in [0,1]^{m_i \times d}. \qquad (4)$$

The normalization can be performed row-wise, i.e. for each node individually, since every node of the graph has exactly one match among the universe points. So for each row (node) $v$, we apply

$$\text{Norm}\left(\hat{F}_i \hat{U}^\top\right)_v := \text{SoftMax}\left((\hat{F}_i \hat{U}^\top)_v\right), \qquad (5)$$

where $\text{SoftMax}(\vec{z})_j = \frac{e^{z_j}}{\sum_k e^{z_k}}$. Note, that in the general case of partial pairwise matching the simple normalization is not applicable, because not every node can be matched.

We predict the discretized object-to-universe matching $\hat{X}_i$ from soft matching $\hat{S}_i$ by solving a linear assignment problem (LAP) with an efficient auction algorithm (Bertsekas 1998)

$$\hat{X}_i := \operatorname*{argmax}_{X \in \mathbb{U}_{m_i d}} \left\langle X, \hat{S}_i \right\rangle = \text{LAP}(\hat{S}_i). \qquad (6)$$

Finally, to obtain cycle-consistent pairwise matchings, we combine corresponding individual object-to-universe matchings, i.e. $\{\hat{X}_{ij} = \hat{X}_i \hat{X}_j^\top\}_{i,j=1}^k$.

For training graph feature encoder $\Psi_\theta$ and universe representations $\hat{U}$, we minimize the discrepancy between soft matchings $\hat{S}_i$ and ground truth universe matchings $X_i^{\text{gt}}$. We

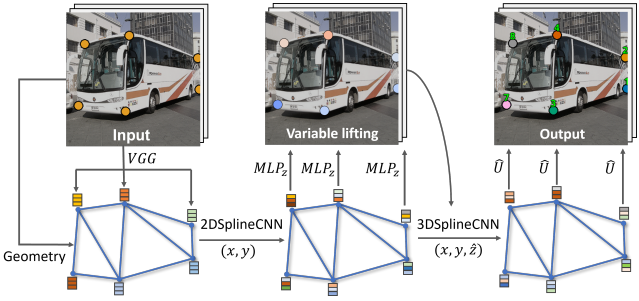

Figure 4: Proposed pipeline for image keypoint matching

propose to use a CrossEntropy loss ($L_{\text{CE}}$) for each node individually

$$L_i := \frac{1}{m_i}\sum_{v=1}^{m_i} L_{\text{CE}}\left((\hat{S}_i)_v, (X_i^{\text{gt}})_v\right), \quad (7)$$

where the $L_{\text{CE}}(\vec{z},\vec{y}\,) = -\sum_j y_j \log z_j$. The total loss is a sum of graph-wise losses, i.e. $L := \sum_{i=1}^{k} L_i$.

**Universe Points Representations** For the fixed node features $\{\hat{F}_i\}_{i=1}^{k}$ and given ground truth object-to-universe matchings $\{X_i^{\text{gt}}\}_{i=1}^{k}$, it is possible to analytically derive the universe points representations, optimal in $L_2$-distance manner, as the average vectors, i.e. *centroids*,

$$U^* := \operatorname*{argmin}_{U} \sum_{i=1}^{k} \left\| X_i^{\text{gt}\top}\hat{F}_i - U \right\|^2 = \frac{1}{k}\sum_{i=1}^{k} X_i^{\text{gt}\top} F_i. \quad (8)$$

The visualization of t-SNE embeddings (in cosine distance) of the learned keypoint features and the universe points representations in Fig. 3 (right plot) shows that the universe points are located at the regions with a high density of the matched keypoints, confirming the above intuition. Note that in the visualization the keypoints are taken from the test set, while universe points are learned from the training set.

**Architecture Design**

In this section, we introduce a neural network architecture for the image keypoint matching problem.

Following previous works (Rolínek et al. 2020), we initialize the node features by interpolating ImageNet pretrained VGG features (Simonyan and Zisserman 2015; Deng et al. 2009) at the keypoint locations (see left plot in Fig. 3), and the edge features – by relative keypoint locations. Further, we assume that the graph structure is fixed by the geometrical properties of keypoints, e.g. by Delaunay triangulation (Delaunay 1934). The graph processing network $\Psi_\theta$ is implemented using SplineCNN (Fey et al. 2018) blocks.

In addition to previous graph neural network architecture from BBGM (Rolínek et al. 2020), we propose to utilize a predicted *third virtual coordinate* $\hat{z}$ in the 3DSplineCNN block. The intuition behind this virtual variable is the nature of keypoints in images, which represent 2D projections of 3D points. The value of the third coordinate is driven by the matching objective.

Overall, our graph processing module $\Psi_\theta$ takes VGG features as input and combines 2DSplineCNN and 3DSplineCNN blocks. The third virtual coordinate $\hat{z}$ is predicted by $\text{MLP}_z$ using the output node features of 2DSplineCNN. The edge features are set as relative keypoint locations. The proposed architecture is illustrated in Fig. 4.

## 4 Experiments

We evaluate the proposed method on various keypoint matching benchmarks. First, we introduce compared methods and metrics. Then we assess different aspects of our method, such as robustness to partiality and scalability, on a controlled synthetic dataset. We evaluate the real-world performance on Pascal VOC and CUB datasets. We also estimate the transferability of the learned node feature encoder to other datasets (Pascal → Willow). Finally, we analyze the contribution of each component of the proposed approach by ablation studies.

**Metrics and Peer Methods**

In *partial* multi-graph matching (i.e. all keypoints are included, $\bigcup$), the pairwise matchings $\{X_{ij} \in \mathbb{P}_{m_i m_j}\}$ contain both inlier matches and non-matches (zero rows/columns in $X_{ij}$). This leads to both types of false positive and false negative errors. To measure the balanced performance of partial matching, we use **F1** score, the harmonic mean of precision and recall. In this setting, we compare our method with recent deep learning-based multi-graph matching methods BBGM-Multi (Rolínek et al. 2020) and JointMGM3D (Ye et al. 2022). We also evaluate learning-free multi-graph matching solvers HiPPI (Bernard et al. 2019b) and MGM-Floyd (Jiang, Wang, and Yan 2020) on Willow dataset.

If graphs contain only (the same number of) inliers (i.e. the intersection of all keypoints, $\bigcap$), the matching problem becomes *full*, and is a special case of partial multi-graph matching. Since in this setting there are no non-matches, we measure the performance of models by computing the accuracy (**Acc**) of predicted full pairwise matchings.

The majority of previous deep graph matching methods consider the full pairwise matching problem (i.e. the intersection of keypoints from two graphs, $\therefore \bigcap \therefore$): GMN (Zanfir and Sminchisescu 2018), SuperGlue (Sarlin et al. 2020), PCA-GM (Wang, Yan, and Yang 2019), LCS (Wang et al. 2020), CIE-H (Yu et al. 2019), GANN-MGM (Wang, Yan, and Yang 2020), DGMC (Fey et al. 2020), BBGM (Rolínek et al. 2020), NHGM-v2 (Wang, Yan, and Yang 2021), GLAM (Liu et al. 2021). Our method can also be evaluated on this restricted setting, as it is a special case of partial multi-matching.

We categorize the methods according to their multi-graph *cycle-consistency* property into 3 categories. First, the methods that consider only two graphs for matching (**no**). Next, the methods that take as input $N$ graphs, and synchronize the matchings between them – the cycle consistency is provided **locally** for the considered graphs. Lastly, the methods that provide cycle consistency without depending on the number of considered input graphs, e.g. by matching each individual graph with a common universe, are categorized as fully cycle-consistent (**yes**).

| Method | F1, ⋃ | Acc, ∴⋂∴ | Cycle |
|---|---|---|---|
| GMN | n/a | 55.3 | no |
| SuperGlue | n/a | 58.1 | no |
| PCA-GM | n/a | 63.8 | no |
| LCS | n/a | 68.5 | no |
| CIE-H | n/a | 68.9 | no |
| DGMC | n/a | 73.0 | no |
| NGM-v2 | n/a | 80.1 | no |
| NHGM-v2 | n/a | 80.4 | no |
| GLAM | n/a | **86.2** | no |
| BBGM | 61.4 | 80.1 | no |
| BBGM-Multi-5 | 62.8 | - | locally |
| JointMGM3D | 42.9 (59.4†) | 59.0 (67.1†) | **yes** |
| URL (**ours**) | **71.7 (82.9†)** | 81.8 (**88.7†**) | **yes** |

Table 1: F1 score (%) of partial multi-graph matching and accuracy of full two-graph matching on the Pascal VOC dataset (averaged over object classes). Since the majority of previous works consider full graph matching, their results are not available (*n/a*) in a partial multi-matching setup.
†Scores in parentheses are directly computed on object-to-universe matchings $\{X_i\}$.

### Synthetic Partial Multi-Matching

We test various aspects of our method on a set of controlled experiments. For this, we create synthetic graphs following the setup from previous works (Wang, Yan, and Yang 2019; Cho, Lee, and Lee 2010). See details in appendix. We enforce *partiality* in generated graphs $\{\mathcal{G}_i\}$ by randomly sampling keypoints among total $N_{\text{univ}}$ universe points for each graph from Binomial distribution with visibility rate $p_{\text{vis}}$, i.e. $m_i \sim \text{Bin}(p_{\text{vis}}, N_{\text{univ}})$.

We compare our method with the strongest peer method BBGM-Multi-$K$. To assess the robustness to partiality, we vary the visibility rate $p_{\text{vis}}$ (with fixed $N_{\text{univ}} = 25$). Further we evaluate scalability and computational efficiency by altering number of universe points $N_{\text{univ}}$ with fixed partiality $p_{\text{vis}} = 0.8$. As demonstrated in Fig. 2, our method scales well to large problems (graphs with up to 1000 nodes) and handles high partiality rates better than the method based on a multi-matching solver. Training of BBGM-Multi-3 on problems with more than 300 universe points does not fit into GPU memory (48 GB).

### Real-World Partial Multi-Matching

To evaluate our method on a real-world problem, we consider multi-image keypoint matching on challenging datasets with natural partiality: the Pascal VOC (Everingham et al. 2010; Bourdev and Malik 2009), and the CUB datasets (Wah et al. 2011). These datasets contain different object classes, and up to 23 universe points depending on object category (Fig. 5). The average visibility rate of Pascal VOC is $p_{\text{vis}}^{\text{VOC}} = 0.72$, and of CUB is $p_{\text{vis}}^{\text{CUB}} = 0.8$. We follow the data pre-processing procedures of previous methods (Wang, Yan, and Yang 2019).

The comparison of matching performance is presented in

| Method | F1, ⋃ | Cycle |
|---|---|---|
| GMN | 79 | no |
| PCA-GM | 83.2 | no |
| CIE-H | 83 | no |
| GANN-MGM | 82.6 | locally |
| URL (**ours**) | **95.1** | **yes** |

Table 2: F1 scores (%) of partial multi-graph matching on CUB dataset. Our method produces the most accurate matchings with built-in cycle consistency.

| Data | Method | Acc, ⋂ | Cycle |
|---|---|---|---|
| — | HiPPI | 88.2 | locally |
| — | MGM-Floyd | 88.3 | locally |
| Pascal → Willow | GMN | 76.2 | no |
| Pascal → Willow | PCA-GM | 85.2 | no |
| Pascal → Willow | CIE-H | 83.5 | no |
| Pascal → Willow | DGMC | 84.1 | no |
| Pascal → Willow | BBGM | 94.5 | no |
| Pascal → Willow | URL-LAP | 95.0 | no |
| Pascal → Willow | URL-HiPPI | **98.9** | locally |

Table 3: Accuracy of full matchings on Willow dataset. Models, pre-trained on (reduced) Pascal VOC dataset, are evaluated on the Willow dataset *without fine-tuning*. Our learned features in URL-HiPPI are better than the hand-crafted ones in a learning-free HiPPI solver, which allows us to also outperform previous learning-based methods.

Tables 1-2 for Pascal VOC and CUB datasets respectively. Per object class F1 scores on Pascal VOC dataset are given in Table 4. So in the real-world setting without any keypoint pre-filtering ( ⋃ ), our method achieves state-of-the-art results on both datasets: 71.7% F1 on Pascal VOC, and 95.1% F1 on CUB. On the object-to-universe matchings, our method achieves an F1-score of 82.9%, and thus outperforms the previous object-to-universe matching method JointMGM3D (Ye et al. 2022) by a large margin. Moreover, our method shows strong performance with 81.8% accuracy on a restricted setting of full two-graph matching ( ∴⋂∴ ). The recent method GLAM (Liu et al. 2021) performs best on this setting. However, in contrast to our method, it learns the graph structures of given points. Thus, the improvement from GLAM is orthogonal to our main contribution.

### Node Feature Transfer

We study the generalization ability of our method by transferring a node feature encoder model $\Psi_\theta$, pre-trained on Pascal VOC dataset (without overlapping images), to Willow dataset (Cho, Alahari, and Ponce 2013). The Willow dataset contains images with consistent object orientations and the same number of keypoints, i.e. all matchings are bijective (see Fig. 5). We evaluate our model by computing cosine similarities of node features with a subsequent discretization by a combinatorial solver. This variation of our model does not use universe points and depends on an external solver (denoted as URL-Solver).

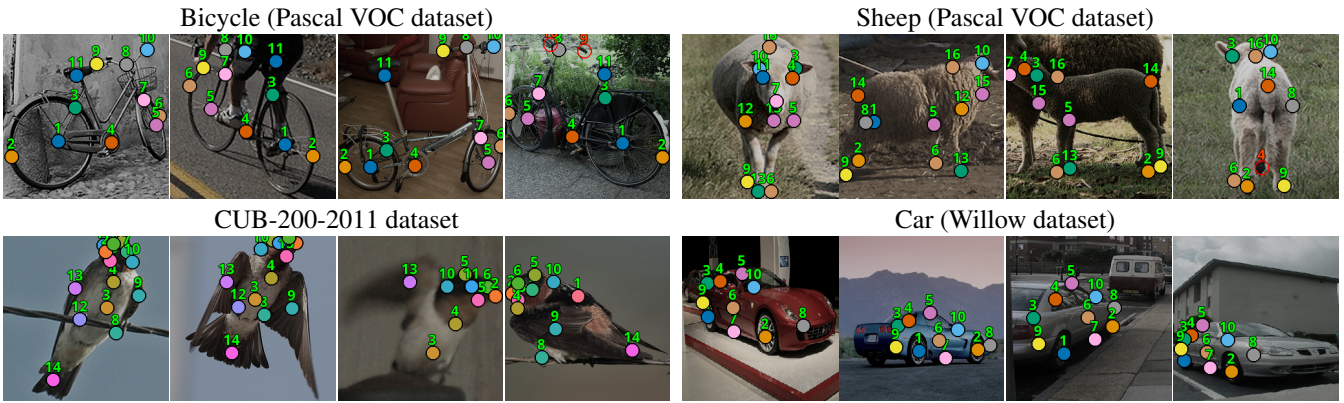

Figure 5: Qualitative results of our method. The *numbers* represent the ground-truth ordering of universe points. Correct predictions are marked with *green* and incorrect ones with *red* numbers. The datasets cover various poses, object deformations, and shape variability of different object classes. The keypoints are matched without any pre-filtering.

| Method | ✈ | 🚲 | 🐦 | 🚤 | 🍾 | 🚌 | 🚗 | 🐈 | 🪑 | 🐄 | 🍽 | 🐕 | 🐎 | 🏍 | 🚶 | 🌱 | 🐑 | 🛋 | 🚆 | 🖥 | Mean F1 |
|---|---|---|---|---|---|---|---|---|---|---|---|---|---|---|---|---|---|---|---|---|---|
| JointMGM3D | 22.9 | 53.3 | 30.6 | 33.6 | 85.7 | 52.3 | 29.6 | 27.7 | 38.7 | 34.9 | 57.8 | 24.4 | 27.7 | 43.9 | 30.6 | 94.3 | 26.7 | 31.9 | 54.0 | 58.6 | 42.9 |
| BBGM-Max | 35.5 | 68.6 | 46.7 | 36.1 | 85.4 | 58.1 | 25.6 | 51.7 | 27.3 | 51.0 | 46.0 | 46.7 | 48.9 | 58.9 | 29.6 | 93.6 | 42.6 | 35.3 | 70.7 | 79.5 | 51.9 |
| BBGM | 42.7 | 70.9 | 57.5 | 46.6 | 85.8 | 64.1 | 51.0 | 63.8 | 42.4 | 63.7 | 47.9 | 61.5 | 63.4 | 69.0 | 46.1 | 94.2 | 57.4 | 39.0 | 78.0 | 82.7 | 61.4 |
| BBGM-Multi-5 | 43.4 | 70.5 | 61.9 | 46.8 | 84.9 | 65.3 | 54.2 | 66.9 | 44.9 | 67.5 | 50.8 | 66.8 | 63.3 | 71.0 | 46.1 | 96.1 | 56.5 | 41.3 | 73.4 | **83.4** | 62.8 |
| URL (**ours**) | **62.7** | **75.2** | **73.0** | **56.7** | **93.7** | **66.2** | **76.7** | **69.2** | **64.9** | **76.6** | 44.6 | **74.4** | **78.8** | **80.9** | **62.5** | **96.9** | **70.3** | **55.4** | 73.6 | 82.1 | **71.7** |

Table 4: F1 scores (%) of partial multi-graph matching on all classes of Pascal VOC dataset (without keypoint filtering, ⋃)

| VGG | 2D | 3D | URL | F1 | Cycle |
|---|---|---|---|---|---|
| ✓ | ✓ | ✗ | ✓ | 67.6±1.0 | yes |
| ✓ | ✗ | ✗ | ✓ | 55.1±0.6 | yes |
| ✓ | ✓ | ✓ | ✗ | 55.4±0.6 | locally |
| ✓ | ✓ | ✓ | ✓ | **71.7**±0.5 | yes |

Table 5: Ablation studies of different components of the proposed method on Pascal VOC dataset. Each building block of the neural network (3D and 2D SplineCNN) significantly contributes to the final performance. The universe representation learning (URL) component has the highest impact both on the F1 score and on the cycle consistency.

From Table 3, we observe that our model URL-HiPPI with a multi-matching solver HiPPI outperforms all other methods, achieving 98.9% matching accuracy. Notably, our approach with the simple linear assignment solver, URL-LAP, produces results comparable with the strong baseline BBGM, which points to the successful transferability of node features in this experiment. The extended results with different training strategies on Pascal VOC and Willow datasets are presented in the supplementary materials.

**Ablation Study**

We evaluate the effectiveness of each component of our proposed method via an ablation study, presented in Table 5. In this experiment, we solve partial multi-matching on the Pascal VOC dataset. We observe that using the 3DSplineCNN block with predicted virtual coordinate gives 4.1% F1-score improvement (rows 1, 4). The graph neural network processing $\Psi_\theta$ (2D and 3D SplineCNN blocks, row 2) corresponds to 16.6% of performance increase, identifying the value of graph-level feature encoding. More importantly, our learned universe points representations (URL, row 3) show a critical influence in providing cycle consistency and in achieving the optimal level of efficacy, accounting for 16.3% gain in F1-score. The reported improvements are statistically significant according to the non-parametric Mann–Whitney U test. The studied methods are run 5 times with random seeds.

## 5 Conclusion

We have presented a deep learning approach for partial multi-graph matching. It builds upon an object-to-universe formulation via learning latent universe point representations. Our important conceptual contribution is that we demonstrate the effectiveness of matching against abstract universe representations, in contrast to pairwise matching, achieving the state-of-the-art performance in practically relevant settings. On a technical level, this is made possible by combining the recent progress in graph-based learning and the architectural improvement with a virtual coordinate lifting. The insights about universe points representations open up new future directions based on exploring unsupervised learning methods for deep multi-graph matching problems. We believe that the observations from our work will be beneficial in other domains with correspondence problems.

# References

Adams, R. P.; and Zemel, R. S. 2011. Ranking via sinkhorn propagation. *arXiv preprint arXiv:1106.1925*.

Bernard, F.; Thunberg, J.; Goncalves, J.; and Theobalt, C. 2019a. Synchronisation of partial multi-matchings via non-negative factorisations. *Pattern Recognition*, 92: 146–155.

Bernard, F.; Thunberg, J.; Swoboda, P.; and Theobalt, C. 2019b. Hippi: Higher-order projected power iterations for scalable multi-matching. In *International Conference on Computer Vision*.

Bertsekas, D. 1998. *Network optimization: continuous and discrete models*. Athena Scientific.

Bourdev, L.; and Malik, J. 2009. Poselets: Body part detectors trained using 3d human pose annotations. In *International Conference on Computer Vision*.

Cho, M.; Alahari, K.; and Ponce, J. 2013. Learning graphs to match. In *International Conference on Computer Vision*.

Cho, M.; Lee, J.; and Lee, K. M. 2010. Reweighted random walks for graph matching. In *European Conference on Computer Vision*.

Delaunay, B. 1934. Sur la sphère vide. A la mémoire de Georges Voronoï. *Bulletin de l'Académie des Sciences de l'URSS. Classe des sciences mathématiques et na*, 6: 793.

Deng, J.; Dong, W.; Socher, R.; Li, L.-J.; Li, K.; and Fei-Fei, L. 2009. ImageNet: A large-scale hierarchical image database. In *Conference on Computer Vision and Pattern Recognition*.

Everingham, M.; Van Gool, L.; Williams, C. K.; Winn, J.; and Zisserman, A. 2010. The pascal visual object classes (voc) challenge. *International Journal of Computer Vision*.

Fey, M.; Lenssen, J. E.; Morris, C.; Masci, J.; and Kriege, N. M. 2020. Deep Graph Matching Consensus. In *International Conference on Logic Programming*.

Fey, M.; Lenssen, J. E.; Weichert, F.; and Müller, H. 2018. Splinecnn: Fast geometric deep learning with continuous b-spline kernels. In *Conference on Computer Vision and Pattern Recognition*.

Gao, M.; Lahner, Z.; Thunberg, J.; Cremers, D.; and Bernard, F. 2021. Isometric Multi-Shape Matching. In *Conference on Computer Vision and Pattern Recognition*.

Huang, Q.-X.; and Guibas, L. 2013. Consistent shape maps via semidefinite programming. In *Computer Graphics Forum*.

Jiang, Z.; Wang, T.; and Yan, J. 2020. Unifying offline and online multi-graph matching via finding shortest paths on supergraph. *IEEE Transactions on Pattern Analysis and Machine Intelligence*.

Kriege, N. M.; Humbeck, L.; and Koch, O. 2019. Chemical Similarity and Substructure Searches. In *Encyclopedia of Bioinformatics and Computational Biology*. Academic Press.

Lawler, E. L. 1963. The quadratic assignment problem. *Management science*, 9(4): 586–599.

Liu, H.; Wang, T.; Li, Y.; Lang, C.; Jin, Y.; and Ling, H. 2021. Joint graph learning and matching for semantic feature correspondence. *arXiv preprint arXiv:2109.00240*.

Munkres, J. 1957. Algorithms for the assignment and transportation problems. *Journal of the society for industrial and applied mathematics*, 5(1): 32–38.

Rolínek, M.; Swoboda, P.; Zietlow, D.; Paulus, A.; Musil, V.; and Martius, G. 2020. Deep graph matching via blackbox differentiation of combinatorial solvers. In *European Conference on Computer Vision*.

Sarlin, P.-E.; DeTone, D.; Malisiewicz, T.; and Rabinovich, A. 2020. Superglue: Learning feature matching with graph neural networks. In *Conference on Computer Vision and Pattern Recognition*.

Scarselli, F.; Gori, M.; Tsoi, A. C.; Hagenbuchner, M.; and Monfardini, G. 2008. The graph neural network model. *IEEE transactions on neural networks*, 20(1): 61–80.

Schmidt, F. R.; Töppe, E.; Cremers, D.; and Boykov, Y. 2007. Intrinsic mean for semi-metrical shape retrieval via graph cuts. In *Joint Pattern Recognition Symposium*.

Simonyan, K.; and Zisserman, A. 2015. Very Deep Convolutional Networks for Large-Scale Image Recognition. In *International Conference on Learning Representations*.

Singh, R.; Xu, J.; and Berger, B. 2008. Global alignment of multiple protein interaction networks with application to functional orthology detection. *Proceedings of the National Academy of Sciences*, 105(35): 12763–12768.

Sun, D.; Yang, X.; Liu, M.-Y.; and Kautz, J. 2018. Pwc-net: Cnns for optical flow using pyramid, warping, and cost volume. In *Conference on Computer Vision and Pattern Recognition*.

Wah, C.; Branson, S.; Welinder, P.; Perona, P.; and Belongie, S. 2011. The caltech-ucsd birds-200-2011 dataset. Technical Report CNS-TR-2011-001, California Institute of Technology.

Wang, L.; Ouyang, W.; Wang, X.; and Lu, H. 2015. Visual tracking with fully convolutional networks. In *International Conference on Computer Vision*.

Wang, R.; Yan, J.; and Yang, X. 2019. Learning combinatorial embedding networks for deep graph matching. In *International Conference on Computer Vision*.

Wang, R.; Yan, J.; and Yang, X. 2020. Graduated Assignment for Joint Multi-Graph Matching and Clustering with Application to Unsupervised Graph Matching Network Learning. In *Conference on Neural Information Processing Systems*.

Wang, R.; Yan, J.; and Yang, X. 2021. Neural graph matching network: Learning lawler's quadratic assignment problem with extension to hypergraph and multiple-graph matching. *IEEE Transactions on Pattern Analysis and Machine Intelligence*.

Wang, T.; Liu, H.; Li, Y.; Jin, Y.; Hou, X.; and Ling, H. 2020. Learning combinatorial solver for graph matching. In *Conference on Computer Vision and Pattern Recognition*.

Ye, Z.; Yenamandra, T.; Bernard, F.; and Cremers, D. 2022. Joint Deep Multi-Graph Matching and 3D Geometry Learning from Inhomogeneous 2D Image Collections. In *Conference on Association for the Advancement of Artificial Intelligencee (AAAI)*.

Yu, T.; Wang, R.; Yan, J.; and Li, B. 2019. Learning deep graph matching with channel-independent embedding and hungarian attention. In *International Conference on Learning Representations*.

Zanfir, A.; and Sminchisescu, C. 2018. Deep learning of graph matching. In *Conference on Computer Vision and Pattern Recognition*.

Zhang, J.; and Philip, S. Y. 2015. Multiple anonymized social networks alignment. In *International Conference on Data Mining*.

# Universe Points Representation Learning for Partial Multi-Graph Matching
## – Supplementary material –

**Zhakshylyk Nurlanov**[1,2], **Frank R. Schmidt**[1], **Florian Bernard**[2]

[1]Bosch Center for Artificial Intelligence
[2]University of Bonn
{zhakshylyk.nurlanov, frank.r.schmidt}@de.bosch.com, fb@uni-bonn.de

## S1  Experimental Details

**Regularization methods**  To avoid over-fitting training data, we use standard regularization techniques. We employ dropout layers (Srivastava et al. 2014) in the network architecture, $L_2$ weight regularization (weight decay), data augmentation, and label smoothing (Müller, Kornblith, and Hinton 2019) in the classification loss. The latter is used since smoothing of a one-hot ground truth vector helps to prevent overconfident predictions of classifiers (Müller, Kornblith, and Hinton 2019). The data augmentations, as implemented by (Buslaev et al. 2020), include geometric transformations, such as in-plane rotations and small perspective transformations, as well as common pixel-wise augmentations.

**Hyperparameters**  We determined the hyperparameters using an automatic hyperparameter optimizer (Bergstra, Yamins, and Cox 2013) on the Pascal VOC dataset and fixed them for all other datasets. Following the protocol of previous works (Zanfir and Sminchisescu 2018; Wang, Yan, and Yang 2020) (according to the authors' code), we find the optimal hyperparameters by the model's performance on a given test set for fairness of comparison with previous methods. The following search intervals were used: learning rate $(10^{-5}, 10^{-1})$, VGG16 backbone learning rate $(10^{-8}, 10^{-1})$ and weight decay $(10^{-8}, 10^{-1})$, which were sampled from log uniform distribution; dropout (0.0, 0.8), label smoothing (0.0, 0.8) and augmentations probability (0.0, 1.0) were drawn from uniform distribution. The optimal hyperparameters that were found and fixed for all experiments are: learning rate $7 \times 10^{-4}$, backbone learning rate $1.527 \times 10^{-4}$, weight decay $3 \times 10^{-7}$, dropout 0.35, label smoothing 0.4, augmentations probability 0.8. The random seed is fixed to 123.

**Keypoint filtering**  As noticed and analyzed in (Rolínek et al. 2020), most previous deep keypoint matching works require keypoint pre-filtering. We introduce both filtered and the more practically relevant unfiltered settings.

The filtered setup assumes an *intersection* between keypoints ($\therefore \bigcap \therefore$) of *two graphs*. That is, only those keypoints that are present in both images are considered as input for determining a matching. As a result, the full matching between the two graphs is expected. In this case, the *accuracy* of the predicted matches is used as an objective metric.

The unfiltered setup considers the *union* of keypoints ($\bigcup$) between *multiple graphs*. In this scenario only a subset of the keypoints between graphs have a match, i.e. we solve partial-to-partial multi-graph matching. In this setting, it is most reasonable to use the F1 score as a performance measure, since it balances precision and recall.

**Runtime**  Training of our method on Pascal VOC dataset takes around 6.5 hours on a single Titan Xp GPU (12 GB). During inference, it processes about 13 images (or 103 keypoints) from the Pascal VOC dataset per second on a laptop-based GeForce GTX 1050 Ti (4 GB).

| Data | Method | Acc, $\bigcap$ | Cycle |
|---|---|---|---|
| — | HiPPI | 88.2 | locally |
| — | MGM-Floyd | 88.3 | locally |
| Pascal → Willow | GMN | 76.2 | no |
| Pascal → Willow | PCA-GM | 85.2 | no |
| Pascal → Willow | CIE-H | 83.5 | no |
| Pascal → Willow | DGMC | 84.1 | no |
| Pascal → Willow | BBGM | 94.5 | no |
| Pascal → Willow | URL-LAP | 95.0 | no |
| Pascal → Willow | URL-HiPPI (ours) | **98.9** | locally |
| Willow | HARG-SSVM | 63.2 | no |
| Willow | LCS | 94.9 | no |
| Willow | DGMC | 97.0 | no |
| Willow | BBGM | 96.8 | no |
| Willow | GANN-MGM | 97.8 | locally |
| Willow | JointMGM3D | 97.8 | yes |
| Willow | NMGM-v2 | **98.2** | locally |
| Willow | URL (ours) | 97.5 | yes |
| Pascal + Willow | GMN | 80.9 | no |
| Pascal + Willow | PCA-GM | 90.2 | no |
| Pascal + Willow | CIE-H | 90.2 | no |
| Pascal + Willow | DGMC | 97.7 | no |
| Pascal + Willow | BBGM | 97.4 | no |
| Pascal + Willow | URL (ours) | **98.3** | yes |

Table S1: Accuracy of full matchings on Willow dataset in different training scenarios.

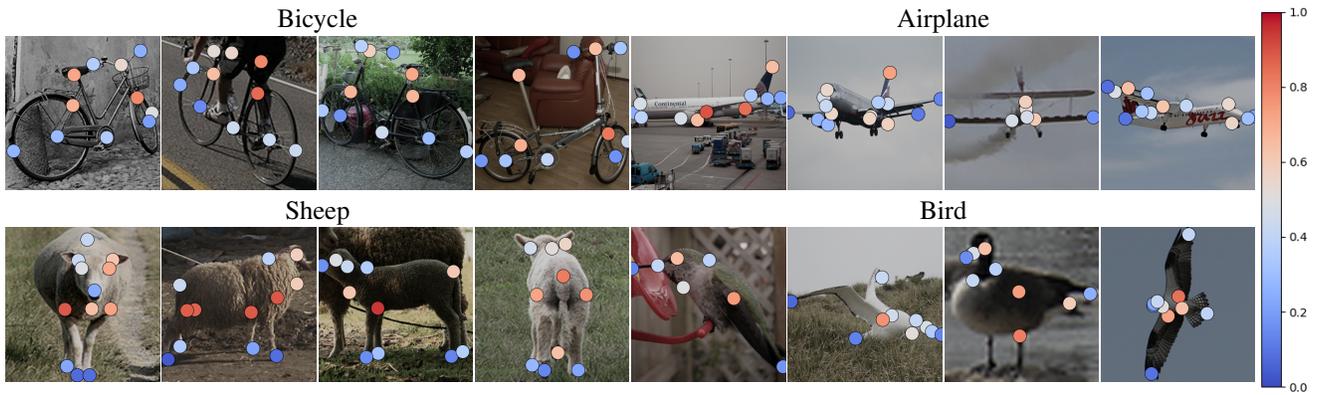

Figure S1: Colormap visualization of the lifted virtual variable for the Pascal VOC dataset (see in color)

**Synthethic dataset**

We create synthetic graphs following the setup from previous works (Wang, Yan, and Yang 2019; Cho, Lee, and Lee 2010). The anchor ground truth graph is generated with a given number of universe points $N_{\text{univ}}$, each with a random 1024-dimensional feature vector from $\mathcal{U}(-1,1)$, and a random 2D coordinate $\sim \mathcal{U}(0, 256)$. The matching graphs for training (200 graphs) and testing (100 graphs) are sampled by perturbing the anchor graph via adding Gaussian noise $\mathcal{N}(0, \sigma_{\text{feat}}^2)$ to features, and applying random affine transform plus additive Gaussian noise $\mathcal{N}(0, \sigma_{\text{coo}}^2)$ to keypoint coordinates. The graph structure is fixed by Delaunay triangulation. Further, we randomly sample $m_i$ keypoints among total $N_{\text{univ}}$ universe points for each graph from Binomial distribution with visibility rate $p_{\text{vis}}$, i.e. $m_i \sim \text{Bin}(p_{\text{vis}}, N_{\text{univ}})$. The default parameters in the synthetic experiments are $\sigma_{\text{feat}} = 1.5, \sigma_{\text{coo}} = 10, p_{\text{vis}} = 0.8, N_{\text{univ}} = 25$.

**Extended results on Willow**

In Table S1 we compare existing methods on three training settings:

- pre-training on reduced Pascal VOC and evaluation on Willow without any fine-tuning (Pascal → Willow);
- training on Willow from scratch (Willow);
- pre-training on reduced Pascal VOC and fine-tuning on Willow (Pascal + Willow).

The Willow dataset is relatively easy for deep graph matching methods due to constant object orientations and the same number of keypoints. Following standard procedure, we use only 20 training samples from the Willow dataset per object class, which makes learning from scratch more challenging.

We observe that when combining pre-trained features from (reduced) Pascal VOC dataset with the multi-matching solver HiPPI (Bernard et al. 2019), we achieve the performance of 98.9% accuracy on the Willow dataset without fine-tuning. In addition to node similarities, the HiPPI method also uses higher-order information via matching the adjacency matrices $\{A_i\}$. The adjacency matrices encode the Gaussian of distances between pairs of points. Hence, for our URL-HiPPI method, we use the third virtual coordinate, together with 2D keypoint locations, to compute the matrices $\{A_i\}$. Overall, our method gives close to perfect matchings in all three training settings.

## S2 Limitations and Future Work

Despite setting the new state of the art on diverse image keypoint matching benchmarks, the proposed URL approach also has limitations.

**Supervision** Similar to other learning-based multi-graph matching approaches, the URL relies on a supervised training strategy, and thus requires the availability of ground truth annotations. We believe that our approach may inspire to conceive training schemes that do not require such annotations, which may for example exploit cycle-consistency as self-supervision.

**Third virtual coordinate** A particular difficulty when matching image keypoints across an image collection is that the keypoints observed in the images represent 2D projections of a (typically unknown) 3D shape. While missing points can adequately be handled in our formulation, to compensate for some of the negative effects due to depth ambiguity, we introduce a *third virtual variable* lifting. In Fig. S1 we show some obtained values of this lifted variable based on a color-coding. While our variable lifting indeed improves matching performance (see Ablation Studies in the main paper), the value of the lifted variable is not interpretable in all cases. Specifically, it does not represent the (relative) depth of the keypoint, but rather some abstract variable. This is expected because the only learning signal it receives is related to matching performance. The incorporation of additional supervisory cues to also predict (relative) depth values is out of the scope of this work, however, it is an interesting direction for future work.

**Object classes** For datasets that comprise multiple object classes, the universe points representations are trained separately for each object class. We leave the exploration of adapting the universe for all classes, e.g. based on conditioning by object class label using an additional object class predictor, for future work.

# References

Bergstra, J.; Yamins, D.; and Cox, D. 2013. Making a science of model search: Hyperparameter optimization in hundreds of dimensions for vision architectures. In *International Conference on Machine Learning*.

Bernard, F.; Thunberg, J.; Swoboda, P.; and Theobalt, C. 2019. Hippi: Higher-order projected power iterations for scalable multi-matching. In *International Conference on Computer Vision*.

Buslaev, A.; Iglovikov, V. I.; Khvedchenya, E.; Parinov, A.; Druzhinin, M.; and Kalinin, A. A. 2020. Albumentations: fast and flexible image augmentations. *Information*, 11(2): 125.

Cho, M.; Lee, J.; and Lee, K. M. 2010. Reweighted random walks for graph matching. In *European Conference on Computer Vision*.

Müller, R.; Kornblith, S.; and Hinton, G. 2019. When does label smoothing help? *arXiv preprint arXiv:1906.02629*.

Rolínek, M.; Swoboda, P.; Zietlow, D.; Paulus, A.; Musil, V.; and Martius, G. 2020. Deep graph matching via blackbox differentiation of combinatorial solvers. In *European Conference on Computer Vision*.

Srivastava, N.; Hinton, G.; Krizhevsky, A.; Sutskever, I.; and Salakhutdinov, R. 2014. Dropout: a simple way to prevent neural networks from overfitting. *The Journal of Machine Learning Research*, 15(1): 1929–1958.

Wang, R.; Yan, J.; and Yang, X. 2019. Learning combinatorial embedding networks for deep graph matching. In *International Conference on Computer Vision*.

Wang, R.; Yan, J.; and Yang, X. 2020. Graduated Assignment for Joint Multi-Graph Matching and Clustering with Application to Unsupervised Graph Matching Network Learning. In *Conference on Neural Information Processing Systems*.

Zanfir, A.; and Sminchisescu, C. 2018. Deep learning of graph matching. In *Conference on Computer Vision and Pattern Recognition*.